\documentclass[10pt,journal,compsoc,twoside]{IEEEtran}
\usepackage[pdftex]{graphicx}
\usepackage{subfigure}
\usepackage{amsmath,amssymb,amsfonts}
\usepackage{color}
\usepackage{blindtext}
\usepackage{algorithm}
\usepackage{algorithmic}
\usepackage{cite}
\usepackage{amsthm}
\usepackage[normalem]{ulem}
\usepackage{url}
\usepackage{comment}
\usepackage{multirow}
\usepackage{xcolor}
\usepackage{diagbox}
\usepackage{indentfirst}

\newtheorem{thm}{Theorem}

\usepackage{hyperref}
\newcounter{ToDo}
\newcounter{gaocomm}
\newcounter{Note}
\definecolor{blue-violet}{rgb}{0.54, 0.17, 0.89}
\definecolor{mygreen}{rgb}{0.0, 0.5, 0.0}
\definecolor{awesome}{rgb}{1.0, 0.13, 0.32}
\definecolor{bostonuniversityred}{rgb}{0.8, 0.0, 0.0}

\begin{document}

\newcommand{\point}{
    \raise0.7ex\hbox{.}
    }

\title{Laplacian LRR on Product Grassmann Manifolds for Human Activity Clustering in Multi-Camera Video Surveillance} 

\author{Boyue~Wang, 
        Yongli~Hu~\IEEEmembership{Member,~IEEE,} Junbin~Gao, Yanfeng~Sun~\IEEEmembership{Member,~IEEE,}
        and Baocai~Yin  ~\IEEEmembership{Member,~IEEE} 
\IEEEcompsocitemizethanks{
\IEEEcompsocthanksitem  Boyue Wang, Yongli Hu and Yanfeng Sun are with Beijing Municipal Key Lab of Multimedia and Intelligent Software Technology, College of Metropolitan Transportation, Beijing University of Technology, Beijing 100124, China. 
E-mail: boyue.wang@emails.bjut.edu.cn, \{huyongli, yfsun\}@bjut.edu.cn
\IEEEcompsocthanksitem Junbin Gao is with the Discipline of Business Analytics, The University of Sydney Business School, The University of Sydney, NSW 2006, Australia. \protect E-mail: junbin.gao@sydney.edu.au 
\IEEEcompsocthanksitem Baocai Yin is with the College of Computer Science and Technology, Faculty of Electronic Information and Electrical Engineering, Dalian University of Technology, Dalian 116620, China; and with Beijing Municipal Key Lab of Multimedia and Intelligent Software Technology at Beijing University of Technology, Beijing 100124, China. \protect E-mail: ybc@bjut.edu.cn
}
}

\markboth{IEEE Transactions on CIRCUITS AND SYSTEMS FOR VIDEO TECHNOLOGY,~Vol.~XX, No.~X, June~2016}%
{Wang \MakeLowercase{\textit{et al.}}: Laplacian LRR on Product Grassmann ...}

\IEEEcompsoctitleabstractindextext{%
\begin{abstract}
In multi-camera video surveillance, it is challenging to represent videos from different cameras properly and fuse them efficiently for specific applications such as human activity recognition and clustering. In this paper, a novel representation for multi-camera video data, namely the Product Grassmann Manifold (PGM), is proposed to model video sequences as points on the Grassmann manifold and integrate them as a whole in the product manifold form. Additionally, with a new geometry metric on the product manifold, the conventional Low Rank Representation (LRR) model is extended onto PGM and the new LRR model can be used for clustering non-linear data, such as multi-camera video data. To evaluate the proposed method, a number of clustering experiments are conducted on several multi-camera video datasets of human activity, including Dongzhimen Transport Hub Crowd action dataset, ACT 42 Human action dataset and SKIG action dataset. The experiment results show that the proposed method outperforms many state-of-the-art clustering methods.
\end{abstract}
}

\maketitle

\section{Introduction}

For the past decades, one has focused on human or crowd activity recognition based on videos, and   significant progresses have been made. However,  most of these works are devoted to  single-camera videos in simple background scenarios \cite{ZhangZhouYaoZhangWangZhang2015,ZhangZhouJiangLi2015,ZhangYaoSunWangZhangLuZhang2014,ZhangYaoSunLu2013,ZhangYaoZhouSunLiu2013,ZhangYaoSunLiu2012}. There exist some natural drawbacks for the single-camera videos-based methods, such as   limited views, objects occlusions, and  low recognition accuracy  under complicated backgrounds. It is difficult to overcome such inherent shortages. In recent years, with the wide use of low-cost cameras in many public places for the purpose of safety, one site is usually covered by several cameras. Therefore, researchers begin to pay attention to human or crowd activity analysis in multi-camera networks, which is meaningful for mitigating the drawbacks of using single-camera mentioned above. Intuitively, the abundant and complementary information from multi-camera systems will improve activity recognition. Towards this goal, many challenges should be overcome, such as how to effectively represent multi-camera data, how to extract the union features from multi-camera videos, how to deal with the discrepancies among different views of videos, and how to fuse information from multiple cameras for the analysis of human or crowd behaviors, and so on.

\begin{figure*}
    \begin{center}
    \includegraphics[width=0.99\textwidth]{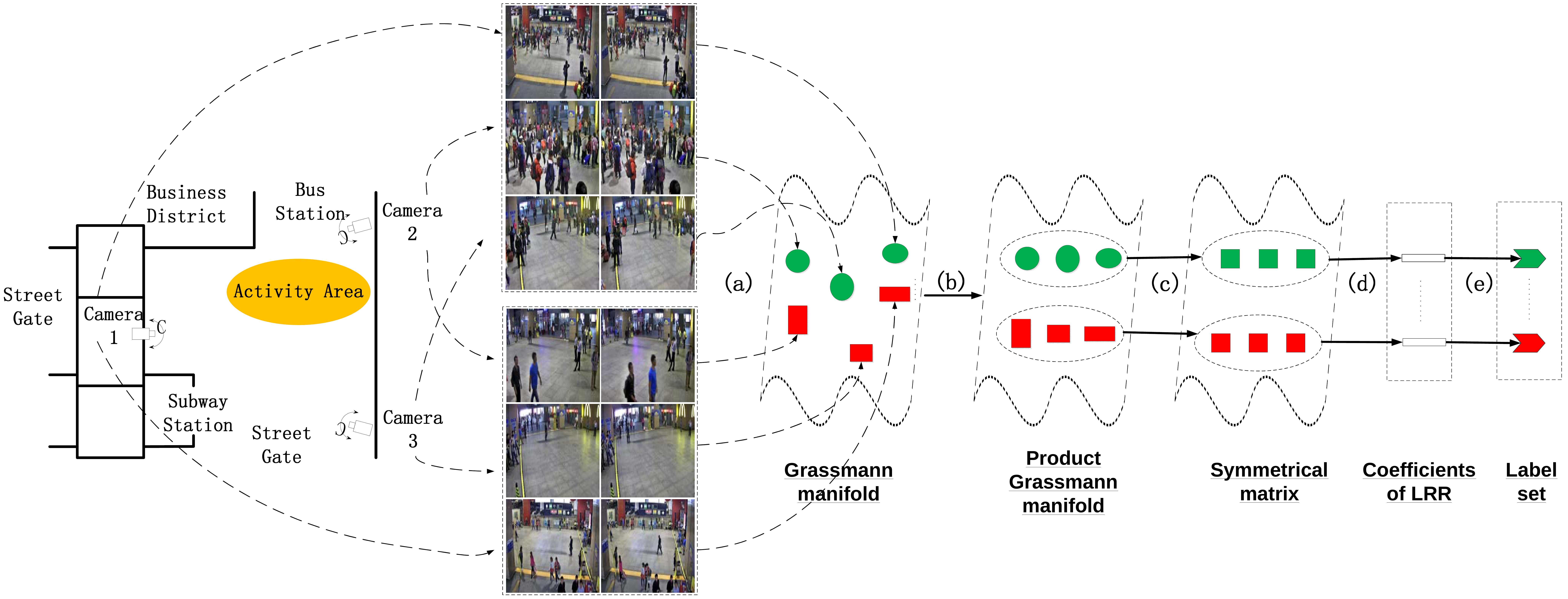}
    \end{center}
    \caption{An overview of our proposed LRR on Product Grassmann manifolds for multi-camera videos clustering. (a) The multi-camera videos are represented as Grassmann points. (b) The Grassmann points from same class are represented as a Product Grassmann point. (c) The Grassmann points are mapped onto symmetrical matrices. (d) The Product Grassmann points are represented by LRR on PGM. (e) Clustering by NCut.}\label{Fig1}
\end{figure*}


 To address these problems, many methods have been proposed to joint videos collected from different cameras and use them in human detection, tracking, and recognition. For human detection, there has been considerable improvement in multi-camera methods compared to single-camera methods  \cite{GehrigDragotti2007,GirodAsronRaneMonedero2005,FlierlVandergheynst2006,GuillemotRoumy2009}. This improvement obviously comes from the fact that an observed human/object in a multi-camera system may appear in different views simultaneously or at different times depending on the overlaps between camera views. Using these multi-information, one can detect objects from each camera's view or combine these information to form a common view for detection. Similarly, the common view can be used for tracking targets  using sequential belief propagation \cite{KimDavis2006,FleuretBerclazLengagneFua2008,DuPiater2007}, where one usually assumes that the topology of camera views is known. Many human action recognition methods are based on human action trajectories extracted from multiple cameras \cite{ShenForoosh2008,ParameswaranChellappa2006} and use the combined trajectories of an object observed in different camera views for activity analysis with the similar approaches developed from single-camera systems. Although these methods can process the dramatic changes in speed and direction of actions, the requirement of accurate tracking trajectories is also challenging. Other approaches like  \cite{LiuYangShah2009,LiuShahKuipersSavarese2011} model as a bag of visual words in each camera view for representing actions, instead of using any tracking information. However this feature representation is sensitive to view changes, and some high level features have to be shared across camera views.

Most multi-camera methods achieve better performance than single-camera methods in many scenarios, yet there still exist some obstacles that need to be conquered. One important issue is how to properly represent the action data captured by multiple cameras, although there are some methods, usually developed for specific applications. The other is how to effectively fuse or combine the information from different cameras as an overall entity.
Majority of existing methods constructs a common map from multiple cameras, which can be fed into any single-camera methods. However this type of  strategies can only be regarded as a na\"{i}ve fusion of multi-camera videos without considering latent relations.





In this paper, we investigate a new way of data representation for multi-camera systems, thus further explore fusion methods for data captured by multiple cameras.
The traditional video features, such as the image bag \cite{ZhangLazebnikSchmid2007} , Local Binary Patterns from Three Orthogonal Plans (LBP-TOP) \cite{ZhaoPietikaeinen2007}  and Improved Dense Trajectories (IDT) \cite{WangSchmid2013} are measured in terms of Euclidean distance.
In fact, it has been proved that many high-dimensional data in computer vision tasks are actually embedded in low dimensional manifolds. Using Euclidean geometry is inappropriate in most of such cases, so,
to get a proper data representation, it is critical to reveal the nonlinear manifold structure underlying these high-dimensional video data.

The classic manifold learning methods are proposed to learn or find nonlinear properties, such as Locally Linear Embedding (LLE) \cite{RoweisSaul2000}, ISOMAP \cite{TenenbaumSilvaLangford2000}, and Locally Linear Projection (LLP) \cite{HeNiyogi2003}.
But these methods only depend on data samples and the manifold underlying the data is unknown. Different from learning nonlinear manifold structure from data, in many scenarios, data are generated from a known manifold. For example, in image analysis, covariance matrices are used to describe region features \cite{TuzelPorikliMeer2006}. A covariance matrix is actually a point on the manifold of symmetric positive definite matrices. Similarly, an image set can be represented as a point on the so-called Grassmann manifold in terms of subspace representation \cite{HarandiSandersonShenLovell2013}.

For the high dimensional video data, to incorporate the possible non-linear intrinsic property, we propose a manifold representation, in which a video is represented as a point on Grassmann manifold. To further fuse the multiple videos, we propose to extend Grassmann manifolds to their product space and obtain a fused representation for multi-camera data, named the Product Grassmann manifold. This is motivated by the fact that product space is good at representing multi-factors determined by multi-subspaces.

To verify the performance of the product manifold representation for multi-camera data, we select human activity clustering in multi-camera surveillance for evaluation. The reason to select clustering tasks is that the clustering problem, especially the video scene clustering, is a challenging problem in computer vision, pattern recognition and signal processing \cite{ElhamifarVidal2013,Vidal2011,XuWunsch-II2005}.
Concretely, we consider the subspace clustering method in our paper. Particularly, the prospective subspace clustering method,  the spectral clustering methods based on affinity matrix, is adopted here.

The main component of the spectral clustering methods is to construct a proper affinity matrix for a dataset and then the affinity matrix is implemented by any clustering algorithms to obtain the final clustering results, such as K-means and Normalized Cuts (NCut) \cite{ShiMalik2000}. There are two classical spectral clustering methods: Sparse Subspace Clustering (SSC) \cite{ElhamifarVidal2013} and low rank representation (LRR) \cite{ LiuLinSunYuMa2013,LiuLinYu2010}. The SSC method assumes that the data of  subspaces are independent and are sparsely represented under the so-called $\ell_1$ Subspace Detection Property \cite{Donoho2004}, in which the within-class affinities are sparse and the between-class affinities are all zeros. It has been proved that under certain conditions the multiple subspace structures can be exactly recovered via $\ell_p  (p\leq 1)$ minimization \cite{LermanZhang2011}. Different from the independent sparse representation for data objects in SSC, the LRR method introduces a holistic constraint, i.e., the low rank or nuclear norm $\|\cdot\|_{*}$ to reveal the latent structural sparse property embedded in the dataset. It has been proven that, when the high-dimensional dataset is actually from a union of several low dimension subspaces, the LRR method can reveal this structure through subspace clustering \cite{LiuLinYu2010}.

Although the subspace clustering methods have good performance in many applications, the current methods assume that data objects come from linear space and the similarity among data is measured in  Euclidean-alike distance. For the manifold representation of multi-camera videos, the clustering methods should be implemented on the manifold. Therefore, we explore the geometry property of the Product Grassmann manifold and extend the conventional LRR method onto Product Grassmann manifold, namely PGLRR model. Furthermore, to capture the local structure of data, we introduce Laplacian constraint to the proposed LRR model on Product Grammann manifold, namely LapPGLRR.

The main idea and framework of the proposed human activity clustering in multi-camera video surveillance based on Laplacian LRR on the product Grassmann manifold is illustrated in Fig. \ref{Fig1}. The contributions of this work are
\begin{itemize}
\item Proposing a new data representation based on the Product Grassmann manifold for multi-camera video data; 
\item Formulating the LRR model on the Product Grassmann Manifold and providing a practical and effective algorithm for the proposed PGLRR model; and
\item Introducing the Laplacian constraint for LRR model on the Product Grassmann manifold.
\end{itemize}

The rest of the paper is organized as follows. In Section \ref{Sec:2}, we review the property of Grassmann manifold and briefly describe the conventional LRR method. In Section \ref{Sec:3}, we propose the Product Grassmann Manifold data representation for multi-camera video data and  present the LRR model on the Product Grassmann Manifold (PGLRR) for clustering. In Section \ref{Sec:4}, we give the solutions to PGLRR and LapPGLRR in detail.  In Section \ref{Sec:5}, the performance of the proposed method is evaluated on clustering problems with several public databases. Finally, the conclusion and the future work are discussed in Section \ref{Sec:6}.

\section{Preliminaries}\label{Sec:2}

\subsection{Grassmann Manifold}
Grassmann manifold $\mathcal{G}(p,d)$ \cite{AbsilMahonySepulchre2008} is the space of all $p$-dimensional linear subspaces of $\mathbb R^d$ for $0\leq p\leq d$. A point on  Grassmann manifold is a $p$-dimensional subspace of $\mathbb R^d$ which can be represented by any orthonormal basis $X=[\mathbf x_1, \mathbf x_2, ..., \mathbf x_p]\in \mathbb R^{d\times p}$. The chosen orthonormal basis is called a representative of its subspace $\text{span}(X)$. Grassmann manifold is an abstract quotient manifold. There are many ways to represent Grassmann manifold. In this paper, we take the way of embedding Grassmann manifold into the space of symmetric matrices $\text{Sym}(d)$.

For convenience, in the sequel we use the same symbol $X$ of the representative orthonormal basis to represent the subspace $\text{span}(X)$. The embedding representation of Grassmann manifold is given by the following mapping \cite{HarandiSandersonShenLovell2013}:

\begin{equation}\label{Grassmann2Sym_mapping}
  \Pi : \mathcal{G}(p,d) \rightarrow \text{Sym}(d), \ \ \ \Pi(X)=XX^T.
\end{equation}
The embedding $\Pi$ is diffeomorphism \cite{HelmkeHuper2007} (a one-to-one, continuous, differentiable mapping with a continuous, differentiable inverse). Given this property, a distance on Grassmann manifold can be induced by the following formula defined by the squared Frobenius norm. Hence it is reasonable to replace the distance on Grassmann manifold with the following distance defined on the symmetric matrix space under this mapping,

\begin{equation}\label{Dist_Grassmann}
 d^2_g(X,Y) = \frac12\|\Pi(X)-\Pi(Y)\|^2_F. 
\end{equation}

\subsection{Product Grassmann Manifold (PGM)}

The PGM is defined as a space of product of multiple Grassmann manifolds, denoted by $\mathcal{PG}_{d:p_1,...,p_M}$. For a given set of natural number $\{p_1, ..., p_M\}$, we  define the PGM $\mathcal{PG}_{d:p_1,...,p_M}$ as the space of $\mathcal{G}(p_1,d)\times \cdot \cdot \cdot \times \mathcal{G}(p_M,d)$.
So a PGM point can be represented as a collection of Grassmannian points, denoted by $[X]=\{{X}^{1},...,{X}^{M}\}$ such that $X^m\in\mathcal{G}(p_m,d),m=1,...,M$.

For our  purpose, 
we consider a weighted sum of Grassmann distances as the distance on PGM,
\begin{equation}\label{Dist_Flag}
  d_\mathcal{PG}([X],[Y])^2 = \sum_{m=1}^M w_m d_g^2(X^m,Y^m),
\end{equation}
where $w_m$ is the weight to represent the importance of the $m$-th Grassmann space. In practice, it can be determined by a data driven manner or according to prior knowledge. In this paper, we simply set all $w_m=1$.
So from \eqref{Dist_Grassmann}, we simply deduce the following distance on PGM,
\begin{equation}\label{Dist_Flag_new}
  d_\mathcal{PG}([X],[Y])^2 = \sum\limits_{m=1}^{M} \frac{1}{2}\|X^{m}(X^{m})^T - Y^{m}(Y^{m})^T \|_F^2.
\end{equation}

\subsection{Low Rank Representation (LRR) \cite{ LiuLinSunYuMa2013}}\label{Subsec:2.1}

Given a set of data drawn from an unknown union of subspaces $\mathbf{X} = [\mathbf x_1, \mathbf x_2, ..., \mathbf x_N]\in\mathbb{R}^{D\times N}$ where $D$ is the data dimension, the objective of subspace clustering is to assign each data sample to its underlying subspace. The basic assumption is that the data in $\mathbf{X}$ are drawn from a collection of $K$ subspaces $\{\mathcal{S}_k\}^K_{k=1}$ of dimensions $\{d_k\}^K_{k=1}$.

According to the principle of self representation of data, each data point from a dataset can be written as a linear combination of the remaining data points, i.e., $\mathbf{X}=\mathbf{X}\mathbf{Z}$, where $\mathbf{Z}\in \mathbb{R}^{N\times N}$ is the coefficient matrix of similarity.

The LRR model is formulated as
\begin{align}
\min_{\mathbf Z, \mathbf E}\|\mathbf E\|^2_{F} + \lambda \|\mathbf Z\|_*,  \text{ s.t. } \mathbf{X} = \mathbf{X}\mathbf{Z}+\mathbf E\label{LRRModel},
\end{align}
where $\mathbf E$ is the error resulting from the self-representation. Similar to the original LRR model, the Frobenius norm can be replaced by the Euclidean $\ell_{2,1}$-norms. LRR takes a holistic view in favor of a coefficient matrix in the lowest rank, measured by the nuclear norm $\|\cdot\|_*$.

 \section{PGM representation of Multi-Camera video Data and Laplacian LRR  Clustering on PGM}\label{Sec:3}
In this section, we first describe the novel representation of the multi-camera video data by PGM and then extend the standard LRR model onto this manifold to obtain a new LRR model on PGM. We also integrate the Laplacian constraint, which captures the local structure of the points on PGM, with the LRR model on PGM to construct a Laplacian LRR model on PGM. Based on the Laplacian LRR model on PGM, we realize the clustering of the multi-camera videos by spectral clustering methods.

\subsection{PGM representation of Multi-Camera video Data}\label{Sec:3.1}
We denote the multi-camera human action video samples by $\mathcal{Y}=\{[Y_1],...,[Y_N]\}$, where $N$ is the number of samples and each sample $[Y_i]$ represents a video set which consists of $M$ video clips of an action captured by $M$ cameras simultaneously, denoted by $[Y_i]=\{C^1_i,...,C_i^M\}$.

For each video clip $C^m_i, m=1,...,M$ in the $i$-th sample $[Y_i]$, {we select all the frames from the clip to form an image set as its delegate, denoted by $\mathbf{S}_i^m,m=1,...,M$. According to \cite{WangHuGaoSunYin2014}, this image set can be represented as a Grassmannian point by using an orthogonal basis of the subspace generated by $\mathbf{S}_i^m$.
Here we adopt the SVD to construct an orthogonal basis, the so-called a Grassmannian point, to represent this video clip, see \cite{WangHuGaoSunYin2014,WangHuGaoSunYin2016} for more details. 
For our purpose in this paper, we give a brief description.
Firstly, we vectorize all frames in $\mathbf{S}_i^m$ and align these vectors as a matrix. For convenience, we still use $\mathbf{S}_i^m$ to denote the matrix. Under the SVD of $\mathbf{S}_i^m$, we construct a $p_m$-dimension subspace  from the first $p_m$ singular vectors. This $p_m$-dimension subspace can be used to approximate the column space of $\mathbf{S}_i^m$. That is, if $\mathbf{S}_i^m=U_i^m\Sigma_i^mV_i^m$ is the SVD, then $X^m_i = U_i^m(:,1:p_m)\in\mathcal{G}(p_m,d)$. $p_m$ could be determined by retaining e.g. $90\%$ of the accumulative eigenvalues of $\Sigma_i^m$.

Combining the $M$ Grassmannian points of the $i$-th sample $[Y_i]$, we  obtain the aforementioned Product Grassmann representation of $[Y_i]$, denoted by $[X_i]=\{X^1_i,...,X^M_i\}\in \mathcal{PG}_{d:p_1,..,p_M}$. By this way, we finally get the PGM representation of the multi-camera human action video samples $\mathcal{Y}=\{[Y_1],...,[Y_N]\}$, denoted by $\mathcal{X}^0=\{ [X_1], [X_2], ..., [X_N] \}$. Next, we will discuss the clustering problem on PGM, i.e. clustering PGM points $[X_i], i=1,...,N$ in $\mathcal{X}^0 $ into their proper classes.

\subsection{LRR on the Product Grassmann Manifolds}


To generalize the LRR model \eqref{LRRModel} onto PGM and implement clustering on the dataset $\mathcal{X}^0$, we first note that in \eqref{LRRModel}
\[
\|\mathbf E\|^2_{F}=\|\mathbf X-\mathbf X\mathbf Z\|^2_{F}=\sum_{i=1}^N \|\mathbf{x}_i-\sum_{j=1}^N \mathbf z_{ij}\mathbf{x}_j\|^2,
\]
where the measure $\|\mathbf{x}_i-\sum_{j=1}^N \mathbf z_{ij}\mathbf{x}_j\|$ is the Euclidean distance between the point $\mathbf x_i$ and its linear combination of all data points including $\mathbf x_i$. Accordingly on PGM we simulate this operation and propose the following form of LRR,
\begin{align}
\begin{aligned}
&\min_{Z}\sum^N_{i=1}\bigg\|[X_i]\ominus (\biguplus^N_{j=1}\mathbf z_{ij}\odot[X_j])\bigg\|_{\mathcal{PG}} + \lambda \|\mathbf Z\|_* , \\
\end{aligned}\label{LRRM}
\end{align}
where $\left\|[X_i]\ominus (\biguplus^N_{j=1}\mathbf z_{ij}\odot[X_j])\right\|_{\mathcal{PG}} $ with the operator $\ominus$ represents the manifold distance between $[X_i]$ and its ``linear'' reconstruction $\biguplus^N_{j=1}\mathbf z_{ij}\odot[X_j]$. Here the combination operators are abstract at this stage.  To get a concrete LRR model on PGM, one needs to define a proper distance and proper combination operations on the manifold.

From the geometric property of the Grassmann manifold, we can use the metric of the Grassmann manifold and the PGM in \eqref{Dist_Grassmann} and \eqref{Dist_Flag} to replace the manifold distance in \eqref{LRRM}, i.e. 
\[
\left\|[X_i]\ominus (\biguplus^N_{j=1}\mathbf z_{ij}\odot[X_j])\right\|_{\mathcal{PG}} =d_\mathcal{PG}([X_i],\biguplus^N_{j=1}\mathbf z_{ij}\odot[X_j]).
\]

In addition,  from \eqref{Grassmann2Sym_mapping} we know that the embedded points in the space of $Sym(d)$ are semi-positive definite matrices. With any positive coefficients, the linear combination on $Sym(d)$ is closed. Thus it is natural to define the ``linearly'' reconstructed Grassmannian point $\biguplus^N_{j=1}\mathbf z_{ij}\odot[X_j]$ as follows,
\[
\Pi \left(\biguplus^N_{j=1}\mathbf z_{ij}\odot[X_j]\right) :=\mathcal{X}\times_4 \mathbf z_i ,
\]
where $\mathbf z_i$ is the $i$-th column of matrix $\mathbf Z$ and
$\mathcal{X}=\{\mathcal{X}_1,\mathcal{X}_2,...,\mathcal{X}_N\}$ is a $4$-th order tensor such that the $4$-th order slices are the $3$rd order tensors $\mathcal{X}_i$, and each  $\mathcal{X}_i$ is constructed by stacking the symmetrically mapped matrices along the 3rd mode. Its mathematical representation is given by

$ \mathcal{X}_i = \{ X_i^1(X_i^1)^T, X_i^2(X_i^2)^T, ..., X_i^M(X_i^M)^T\}\subset
\text{Sym}(d)$.
And $\times_4$ means the mode-4 multiplication  of a tensor and a vector (and/or a matrix) \cite{KoldaBader2009}.

Finally, we can construct the LRR model on the PGM followed as \cite{KoldaBader2009}

\begin{equation}\label{flaglrr}
\min\limits_{\mathbf{E},\mathbf Z}\|\mathbf Z\|_* + \lambda\|\mathbf{E}\|_F^2 \ \ \  \text{s.t.} \ \ \ \mathcal{X}=\mathcal{X}\times_4\mathbf Z+\mathbf{E}.
\end{equation}
In other words, the LRR on PGM is implemented on the product of the symmetric matrix spaces.

\subsection{Laplacian LRR on The Product Grassmann Manifolds}
The low rank term in the LRR model \eqref{flaglrr} on PGM makes a holistic constraint on the  coefficient matrix $\mathbf Z$. However, the points on PGM also have their geometric property in sense of geodesic distance on the manifold. So this geometric property should also be converted to their corresponding LRR representation coefficient matrix $\mathbf Z$. From this observation, we further add a geometric constraint on the coefficient matrix $\mathbf Z$ in terms of Laplacian Matrix to get the following Laplacian LRR on the Product Grassmann Manifolds.

For the coefficient matrix $\mathbf Z$, we consider imposing the local geometrical structures to enforce the coefficient matrix preserving the intrinsic structures of original data on the manifold. Under the LRR model, $\mathbf z_i$ and $\mathbf z_j$ are the new representations of data objects $\mathbf x_i$ and $\mathbf x_j$, respectively. The distance between $\mathbf z_i$ and $\mathbf z_j$ defines certain similarity between data $\mathbf x_i$ and $\mathbf x_j$. Laplacian regularization is considered as a good way to preserve the similarity. As a result, we add the Laplacian regularization into the  objective function \eqref{flaglrr} as follows,
\begin{equation}\label{LapPGlrr-1}
\begin{aligned}
&\min\limits_{\mathbf Z,\mathbf E}\|\mathbf Z\|_* + \lambda\|\mathbf{E}\|_F^2 + \beta\sum\limits_{i,j}\|\mathbf z_i-\mathbf z_j\|_2^2 w_{ij} \\ &\text{s.t.} \ \ \mathcal{X} = \mathcal{X}_{\times_4}\mathbf Z + \mathbf{E}
\end{aligned}
\end{equation}
where $w_{ij}$ is the geodesic distance between the Product Grassmannian points $[X_i]$ and $[X_j]$. The simplified form of the $3$rd term in \eqref{LapPGlrr-1}  is given by
\begin{equation}
\begin{aligned}
&\sum\limits_{i,j}\|\mathbf z_i- \mathbf z_j\|_2^2 w_{ij} 
= 2\text{tr}(\mathbf Z^T L \mathbf Z),
\end{aligned}
\end{equation}
where $L=D-W$ and $D$ is the diagonal matrix with diagonal elements $d_{ii}=\sum\limits_{j}w_{ij}$. The element $w_{ij}$ is defined by the geodesic distance, refer to \eqref{Dist_Flag_new}
\[
w_{ij} = d_\mathcal{PG}([X_i],[X_j]) = \sqrt{\sum^M_{m=1}d^2_g(X^m_i, X^m_j)}.
\]
Thus, the objective function \eqref{LapPGlrr-1} can be rewritten as,
\begin{equation}\label{LapPGlrr-2}
\begin{aligned}
&\min\limits_{\mathbf{E},\mathbf Z}\|\mathbf Z\|_* + \lambda\|\mathbf{E}\|_F^2 + 2\beta \text{tr}(\mathbf ZL\mathbf Z^T), \ \\ &\text{s.t.} \ \ \mathcal{X} = \mathcal{X}_{\times_4}\mathbf Z + \mathbf{E}.
\end{aligned}
\end{equation}
For convenience,we abbreviate it by LapPGLRR.

\subsection{Clustering algorithm for the multi-camera videos by the Laplacian LRR on PGM}
For the multi-camera videos $\mathcal{Y} = \{[Y_1], ..., [Y_N]\}$, we first find out their PGM representation $\mathcal{X}^0 = \{[X_1], ..., [X_N]\}$ as described in Section 3.1. After formulating the PGLRR model in \eqref{flaglrr} or LapPGLRR model in \eqref{LapPGlrr-2} and solving these optimization problems (How to solve these problems will be discussed in the next Section.), we can find the data representation coefficient matrix $\mathbf Z$. Under the data self-representative principle used in the models, the element $\mathbf z_{ij}\in \mathbf Z$ represents the similarity between data $i$ and $j$.  So a natural way is to define the affinity matrix $\hat{\mathbf Z}=(|\mathbf Z|+|\mathbf Z^T|)/2$ for model \eqref{flaglrr} or \eqref{LapPGlrr-2}. This affinity matrix $\hat{\mathbf Z}$ can be performed on any spectral clustering algorithms, such as Ncut \cite{ShiMalik2000}, to obtain the final clustering result. The whole clustering procedure of the proposed clustering algorithm for multi-camera videos by the Laplacian LRR on PGM is summarized as Algorithm~\ref{Alg1}.

\begin{algorithm}
\renewcommand{\algorithmicrequire}{\textbf{Input:}}
\renewcommand\algorithmicensure {\textbf{Output:} }
\caption{Clustering algorithm for multi-camera videos by the Laplacian LRR on PGM}\label{Alg1}
\begin{algorithmic}[1]
\REQUIRE The multi-camera videos for clustering $\mathcal Y$.  \\
\ENSURE  The clustering results of $\mathcal Y$.\\
\STATE   Representing $\mathcal Y$ as PGM pionts $\mathcal{X}^0$ as Section \ref{Sec:3.1};
\STATE   Calculating geodesic distance $w_{ij}$ between $[X_i]$ and $[X_j]$ and constructing the Laplace matrix $L$;
\STATE   Obtaining the LRR representation $\mathbf Z$ of $\mathcal{X}$ by \eqref{LapPGlrr-2} ;
\STATE   Computing the affinity matrix $\hat{\mathbf Z}=(|\mathbf Z|+|\mathbf Z^T|)/2$;
\STATE   Implementing NCut$(\hat{\mathbf Z})$ to get the final clustering result of $\mathcal Y$.
\end{algorithmic}
\end{algorithm}

\section{Solution to Laplacian LRR on PGM}\label{Sec:4}
First, we give the solution to the LRR on PGM in \eqref{flaglrr} in which only the holistic low rank constraint is considered. Then the solution to the LaplacianLRR on PGM in \eqref{LapPGlrr-2} is discussed.
\subsection{Solution to LRR on PGM}
To avoid tedious calculation between the $4$-order tensor and the matrix in \eqref{flaglrr}, we briefly analyze the representation of the reconstruction tensor error $\mathbf{E}$ and translate the optimization problem into an equivalent and solvable optimization model.

The explicit form of $\|\textbf E\|_F^2$ is given by
\begin{align*}
 \|\mathbf{E}\|_F^2 &= \sum^N_{i=1}\sum^M_{m=1}\|(X_i^m(X_i^m)^T-\sum\limits_{j=1}^{N}\mathbf{z}_{ij}(X_j^m(X_j^m)^T))\|_F^2. 
\end{align*}
To simplify the expression for $\|\mathbf{E}\|_F^2$, we firstly note that the matrix property
\[
\|A\|_F^2 = \text{tr}(A^{T}A)
\]
and denote
\begin{equation}\label{Delta_ijmn}
\Delta_{ij}^m = \text{tr}[((X_j^m)^{T}X_i^m)((X_i^m)^{T}X_j^m)].
\end{equation}

Observing that $\Delta_{ij}^m=\Delta_{ji}^m$, we define $M$ $N \times N$ symmetric matrices as
\begin{equation}\label{Delta_mn}
\Delta^{m} = (\Delta_{ij}^m)_{i=1,j=1}^{N},  \ \ \ m = 1, 2, ..., M.
\end{equation}
Moreover, it is easy to prove that
\begin{align}
\|\mathbf{E}\|_F^2 = -2\text{tr}(\mathbf Z\Delta)+\text{tr}(\mathbf Z\Delta \mathbf Z^T) + \text{const}, \label{EReconstruct}
\end{align}
where $\Delta=\sum_{m=1}^{M}\Delta^{m}$ and  the term const collects all the  terms irrelevant to the variable $\mathbf Z$.

Similar to \cite{WangHuGaoSunYin2014}, it is easy to  prove that $\Delta$ is positive semi-definite. Consequently, we have a spectral decomposition of $\Delta$ given by
\[
\Delta  = U D U^T,
\]
where $U^TU = I$ and $D = \text{diag}(\sigma_i)$ with non-negative eigenvalues $\sigma_i$. So   \eqref{EReconstruct} becomes
\[
\|\mathbf E\|_F^2  = \|\mathbf Z\Delta^{\frac{1}{2}}-\Delta^{\frac{1}{2}}\|_F^2 + \text{const},
\]
after variable elimination and problem \eqref{flaglrr} can be converted to
\begin{equation}\label{FLRR_Final}
\begin{aligned}
\min\limits_{\mathbf Z}\|\mathbf Z\Delta^{\frac{1}{2}}-\Delta^{\frac{1}{2}}\|_F^2 + \lambda\|\mathbf Z\|_*.
\end{aligned}
\end{equation}

There exists a closed form solution to the optimization problem \eqref{FLRR_Final} following \cite{FavaroVidalRavichandran2011}, and it is given by  the following theorem.
\begin{thm}\label{thm_1}
Given that $\Delta = UDU^T$ as defined above, the solution to \eqref{FLRR_Final} is given by
\[
\mathbf{Z}^* = U D_{\lambda}U^T,
\]
where $D_{\lambda}$ is a diagonal matrix with its $i$-th element defined by
\[
D_{\lambda}(i,i) = \begin{cases} 1 - \frac{\lambda}{\sigma_i}  & \text{ if } \sigma_i > \lambda, \\
0 & \text{ otherwise}.
\end{cases}
\]
\end{thm}

\subsection{Solution to Laplacian LRR on PGM}

By using the same technique deriving  the above PGLRR algorithm, it is easy to deduce the equivalent form of the model in \eqref{LapPGlrr-2} as follows,
\begin{equation}\label{LapPGlrr-3}
\begin{aligned}
\min\limits_{\mathbf Z} - 2\lambda\text{tr}(\mathbf Z\Delta) + \lambda\text{tr}(\mathbf Z\Delta \mathbf Z^T) + 2\beta\text{tr}(\mathbf ZL\mathbf Z^T) + \|\mathbf Z\|_*.
\end{aligned}
\end{equation}

We employ the Augmented Lagrangian Multiplier (ALM) to solve this problem. So we let $J=\mathbf Z$ to separate the variable $\textbf Z$ from different terms. Then  problem \eqref{LapPGlrr-3} can be formulated as follows,
\begin{equation}
\begin{aligned}
&\min\limits_{\mathbf Z,J} - 2\lambda\text{tr}(\mathbf  Z\Delta) + \lambda\text{tr}(\mathbf Z\Delta \mathbf Z^T) + 2\beta\text{tr}(\mathbf ZL\mathbf Z^T) + \|J\|_* \\ &\text{s.t.} \ \ J=\mathbf Z
\end{aligned}
\end{equation}

Its Augmented Lagrangian Multiplier formulation can be defined as the following unconstrained optimization,
\begin{equation}
\begin{aligned}
& - 2\lambda\text{tr}(\mathbf Z\Delta) + \lambda\text{tr}(\mathbf Z\Delta \mathbf Z^T) + 2\beta\text{tr}(\mathbf ZL\mathbf Z^T) + \|J\|_* \\
  &+ \langle A,\mathbf Z-J \rangle + \frac{\mu}{2}\|\mathbf Z-J\|_F^2
\end{aligned}
\end{equation}
where $A$ is the Lagrangian Multiplier and $\mu$ is a weight to tune the error term of $\|\mathbf Z-J\|_F^2$.

Now, the above problem can be solved by solving the following two subproblems in an alternative manner, fixing $\mathbf Z$ or $J$ to optimize the other, respectively.

When fixing $\mathbf Z$, the following subproblem is solved to update $J$
\begin{equation}\label{subproblemJ}
\begin{aligned}
\min\limits_J \|J\|_* + \langle A,\mathbf Z-J \rangle + \frac{\mu}{2}\|\mathbf Z-J\|_F^2
\end{aligned}
\end{equation}

When fixing $J$, the following subproblem is solved to update $\mathbf Z$
\begin{equation}\label{subproblemZ}
\begin{aligned}
\min\limits_{\mathbf Z} & - 2\lambda\text{tr}(\mathbf Z\Delta) + \lambda\text{tr}(\mathbf Z\Delta \mathbf Z^T) + 2\beta\text{tr}(\mathbf ZL\mathbf Z^T) \\
     &+ \langle A,\mathbf Z-J \rangle + \frac{\mu}{2}\|\mathbf Z-J\|_F^2
\end{aligned}
\end{equation}

Subproblem  \eqref{subproblemJ} can be solved by the following way. Firstly, the optimization is revised as follows,
\begin{equation}\label{requireJ}
  \begin{split}
      \min\limits_{J}(||J||_*+\frac{\mu}{2}||J-(\mathbf Z+\frac{A}{\mu})||_F^2).
  \end{split}
\end{equation}
\eqref{requireJ} has a closed-form solution given by,
\[
J^* = \Theta_{\mu^{-1}}(\mathbf Z + \frac{A}{\mu}),
\]
where $\Theta(\cdot)$ denotes the singular value thresholding operator (SVT), see \cite{CaiCandesShen2008}.

The subproblem in \eqref{subproblemZ} is a quadratic optimization problem with respect to $\mathbf Z$. The closed-form solution is given by
\begin{equation}
\begin{aligned}
\mathbf Z = (2\lambda\Delta + \mu J - A)(2\lambda\Delta + 2\beta L + \lambda I)^{-1}
\end{aligned}
\end{equation}

Solving the above two subproblems alternatively results in the complete solution to LapPGLRR. The whole procedure of LapPGLRR through solving problem \eqref{LapPGlrr-3} is summarized in Algorithm \ref{wholeAlg}.

\begin{algorithm}
\renewcommand{\algorithmicrequire}{\textbf{Input:}}
\renewcommand\algorithmicensure {\textbf{Output:} }
\caption{ Solving Problem \eqref{LapPGlrr-3} by ALM.}\label{wholeAlg}
\begin{algorithmic}[1]
\REQUIRE The Product Grassmann sample set $\{[X_i]\}_{i=1}^N$, $[X_i]\in \mathcal{PG}_{n:p_1,..,p_M}$, and the balancing parameters $\lambda$ and $\beta$. \\
\ENSURE  The Low-Rank Representation $Z$ ~~\\
\STATE   Initialize:$J=Z=0,A=0,\mu=10^{-6},\mu_{max}=10^{10}$ and $\varepsilon=10^{-8}$
\FOR{m=1:M}
\FOR{i=1:N}
\FOR{j=1:N}
\STATE   $\Delta_{ij}^m\leftarrow \mbox{tr}[({X_j^m}^{T}X_i^m)({X_i^m}^{T}X_j^m)]$;
\ENDFOR
\ENDFOR
\ENDFOR
\FOR{m=1:M}
\STATE $\Delta \leftarrow \Delta + \Delta^m_{::}$;
\ENDFOR
\WHILE   {not converged}
\STATE   fix $Z$ and update $J$ by \\$J\leftarrow \min\limits_{J}(||J||_*+\langle A,Z-J\rangle+\frac{\mu}{2}||Z-J||_F^2) $;
\STATE   fix $J$ and update $Z$ by \\$Z = (2\lambda\Delta + \mu J - A)(2\lambda\Delta + 2\beta L + \lambda I)^{-1}$ ;
\STATE   update the multipliers: \\ $A\leftarrow A+\mu(Z-J)$
\STATE   update the parameter $\mu$ by \\$\mu \leftarrow \min(\rho\mu,\mu_{\mbox{max}})$
\STATE   check the convergence condition: \\ {$\|Z-J\|_\infty <\varepsilon$}
\ENDWHILE
\end{algorithmic}
\end{algorithm}

\section{Experiments}\label{Sec:5}
In this section, we evaluate the performance of our proposed clustering approaches on a human activity multi-camera video dataset we collected, \emph{the Dongzhimen Transport Hub Crowd Dataset}; and other two multi-view or multi-modality individual action datasets, \emph{the ACT42 action dataset}\footnote{\url{http://vipl.ict.ac.cn/rgbd-action-dataset/download}.} and the \emph{SKIG action clips dataset}\footnote{\url{http://lshao.staff.shef.ac.uk/data/SheffieldKinectGesture.htm}.}. The experiments are conducted on these three datasets with four state-of-the-art manifold-based clustering methods and three classic clustering methods using both LBP-TOP video features  \cite{ZhaoPietikaeinen2007} and IDT video features \cite{WangSchmid2013}. The IDT video feature is one of the state-of-the-art effective representations for action recognition in videos. For a summary, we list all the comparing methods as follows:


\begin{itemize}
  \item \textbf{GLRR-F} \cite{WangHuGaoSunYin2014}: Low Rank Representation on Grassmann Manifold embeds the image sets into the Grassmann manifold and extends the standard LRR model onto the Grassmann manifold.
  \item \textbf{SCGSM} \cite{TuragaVeeraraghavanSrivastavaChellappa2011}: Statistical computations on the Grassmann and Stiefel manifolds uses a statistical model derived from Riemannian geometry of the manifold.
  \item \textbf{SMCE} \cite{ElhamifarVidalNips2011}: Sparse Manifold Clustering and the Embedding utilizes the local manifold structure to find a small neighborhood around each data point and connects each point to its neighbours with appropriate weights.
  \item \textbf{LS3C} \cite{PatelNguyenVidal2013}: Latent Space Sparse Subspace Clustering learns the projection of data and finds the sparse coefficients in the low-dimensional latent space.
  \item \textbf{LRR+IDT/LBP-TOP}: The standard LRR method \cite{LiuLinSunYuMa2013} is implemented with the IDT features or LBP-TOP features of videos instead of the raw data.
  \item \textbf{K-means+IDT/LBP-TOP}: K-means algorithm is implemented  on the IDT features or LBP-TOP features of videos.
  \item \textbf{SPC+IDT/LBP-TOP}: Spectral Clustering method \cite{NgJordanWeiss2002} is implemented on the IDT features or LBP-TOP features of videos.
\end{itemize}

\begin{figure}
    \begin{center}
    \includegraphics[width=0.45\textwidth]{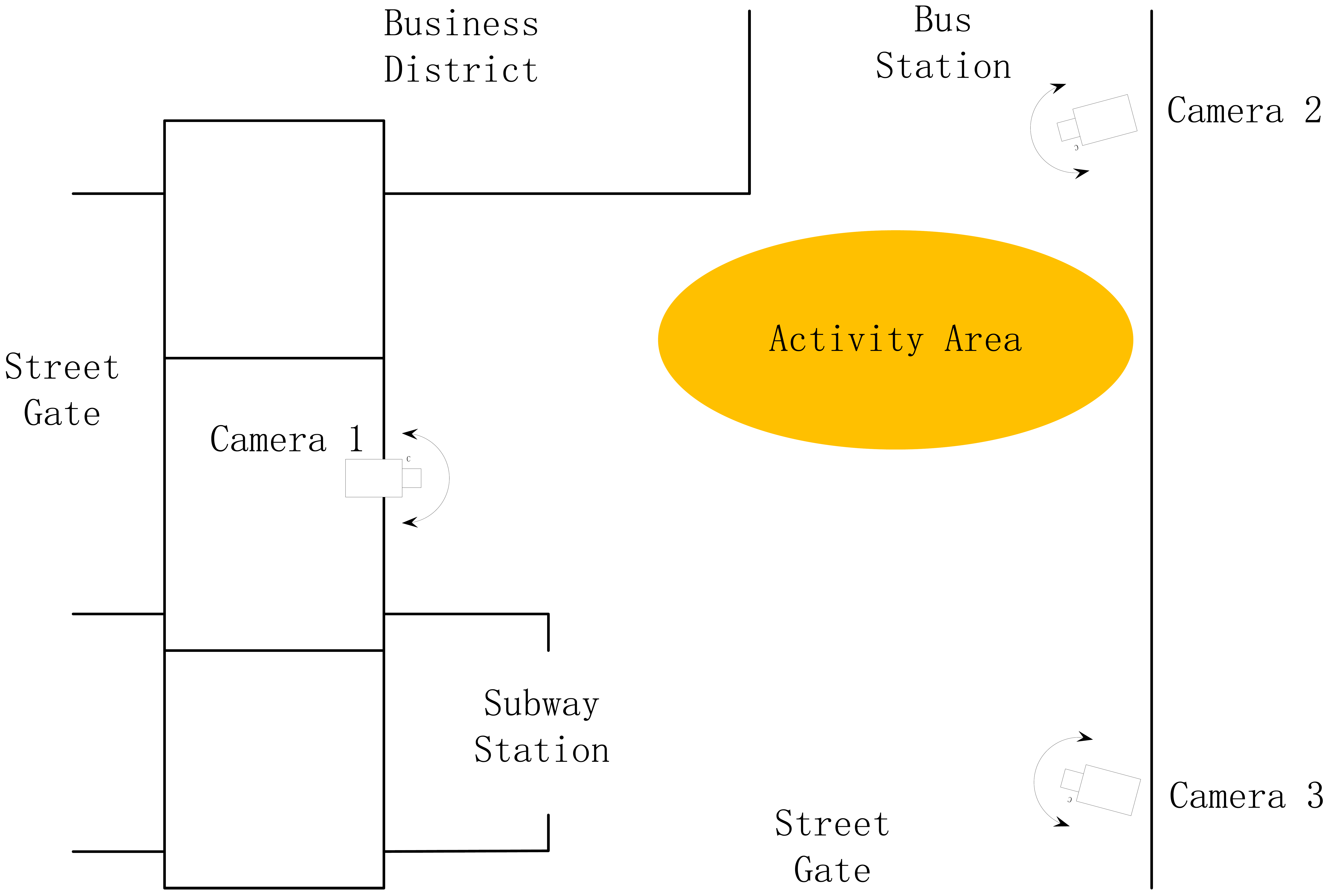}
    \end{center}
    \caption{An illustrative example for the positions of three cameras in Dongzhimen Transport Hub.}\label{Fig2}
\end{figure}

In all the above methods, Grassmann manifold representation, IDT video features and LBP-TOP video features are extracted from videos. In IDT, one firstly learns a codebook from a set of motion trajectory features constructed by several descriptors (i.e., Trajectory, HOG, HOF and MBF). However, at the next encoding stage, some information may be lost when a trajectory feature is represented by the nearest visual word in codebook. LBP-TOP describes videos using three Histograms of Spatial-temporal LBP features, which captures local spatial and temporal information at pixel level but fails to retain the global structure relation when computing the histograms. Grassmann manifold representation is generated by selecting the first $p$ singular vectors according to the descending eigenvalues in SVD, which can preserve the principal information of videos. Through the following experiments, we will prove the advantages of using Grassmann manifold representation. 

Our proposed method PGLRR consists of three main ingredients: Product manifold, Grassmann manifold representation and LRR. In the following experiments, we intend to find out which one of them plays a key role in boosting clustering accuracy.  We take three compared methods, GLRR-F, SCGSM and LRR+IDT/LBP-TOP, as the most important baselines. PGLRR extends GLRR-F onto product manifold, thus comparing  the performance of PGLRR with GLRR-F will demonstrate the importance of product manifold.
To evaluate the importance of Grassmann manifold representation, we compare PGLRR's performance to SCGSM's without LRR. Similarly, for assessing the impact of LRR, we can compare PGLRR's performance to LRR+IDT/LBP-TOP's without Grassmann manifold representation. We conclude that all three ingredients improve the effectiveness of PGLRR, respectively, as demonstrated in the following experiments.

\subsection{Experimental Datasets}

\textbf{1) Dongzhimen Transport Hub Crowd Dataset (DTHC)}

We construct a multi-camera human activity dataset to evaluate the proposed methods. We choose the Dongzhimen Transport Hub in Beijing, China, as a site to collect multi-camera data.  Dongzhimen Transport Hub is one of the busiest transport hubs in Beijing. Many passengers take transfer between different routes in this hub everyday, hence there exist complicated crowd activities. We deploy three cameras in a hall (as shown in  Fig.~\ref{Fig2}) to capture the videos of passengers. The dataset is captured from 06:00 to 22:00 on a Saturday. We pick up  182 multi-camera samples as our experimental data. Each sample has three video clips. Samples of this dataset  are labeled with three level of crowd actions: heavy, light, and medium. There are  48 samples of heavy level, 76 samples of light level and 58 samples of medium level. The frames are converted to gray images and each image is normalized to size $32 \times 58$. Some samples of the Dongzhimen Transport Hub Crowd dataset are shown in Fig.~\ref{DZMfig}.\footnote{We will make the Dongzhimen Transport Hub Crowd Dataset public soon.}
\begin{figure*}[!th]
\centering
\subfigure[]{ \label{SKIGfig:a} 
\includegraphics[width=0.87\textwidth]{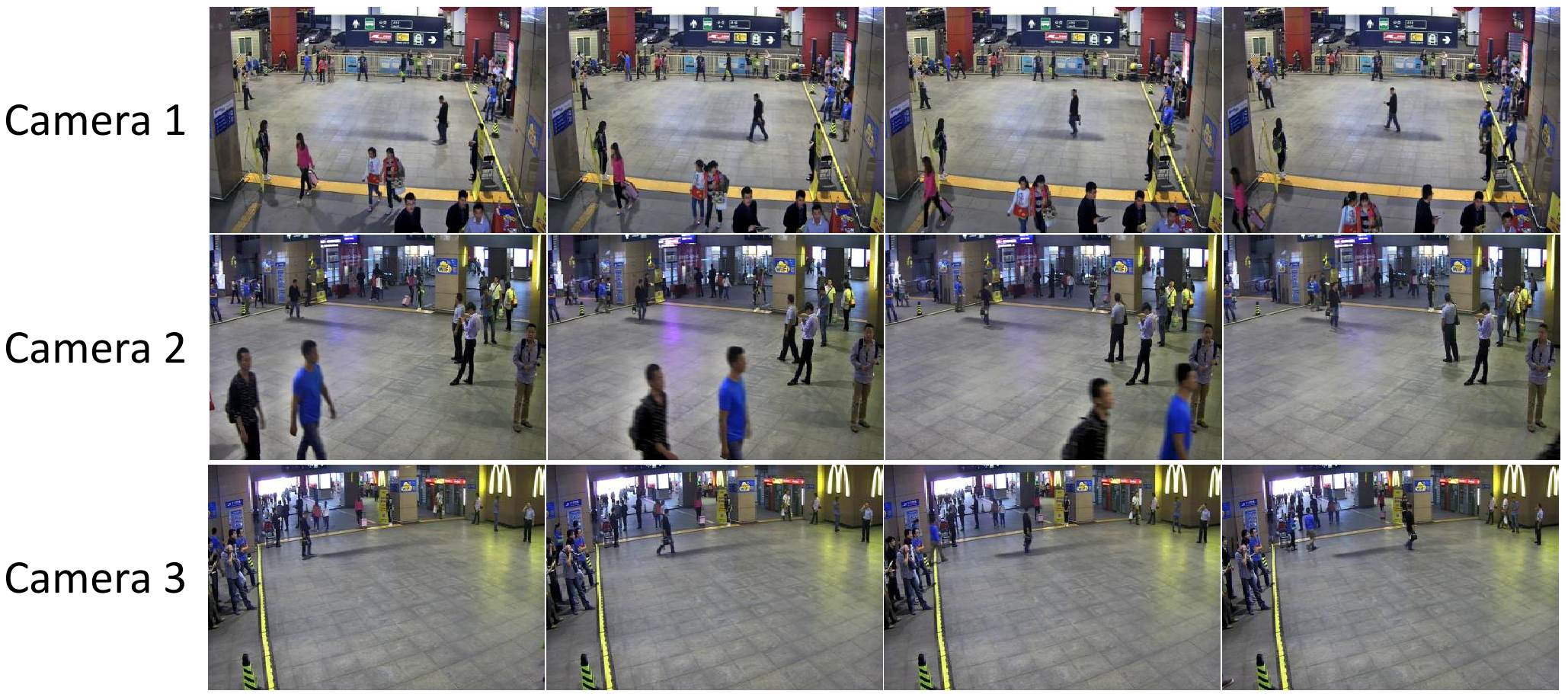}}
\hspace{0.1in}
\subfigure[]{ \label{SKIGfig:b} 
\includegraphics[width=0.87\textwidth]{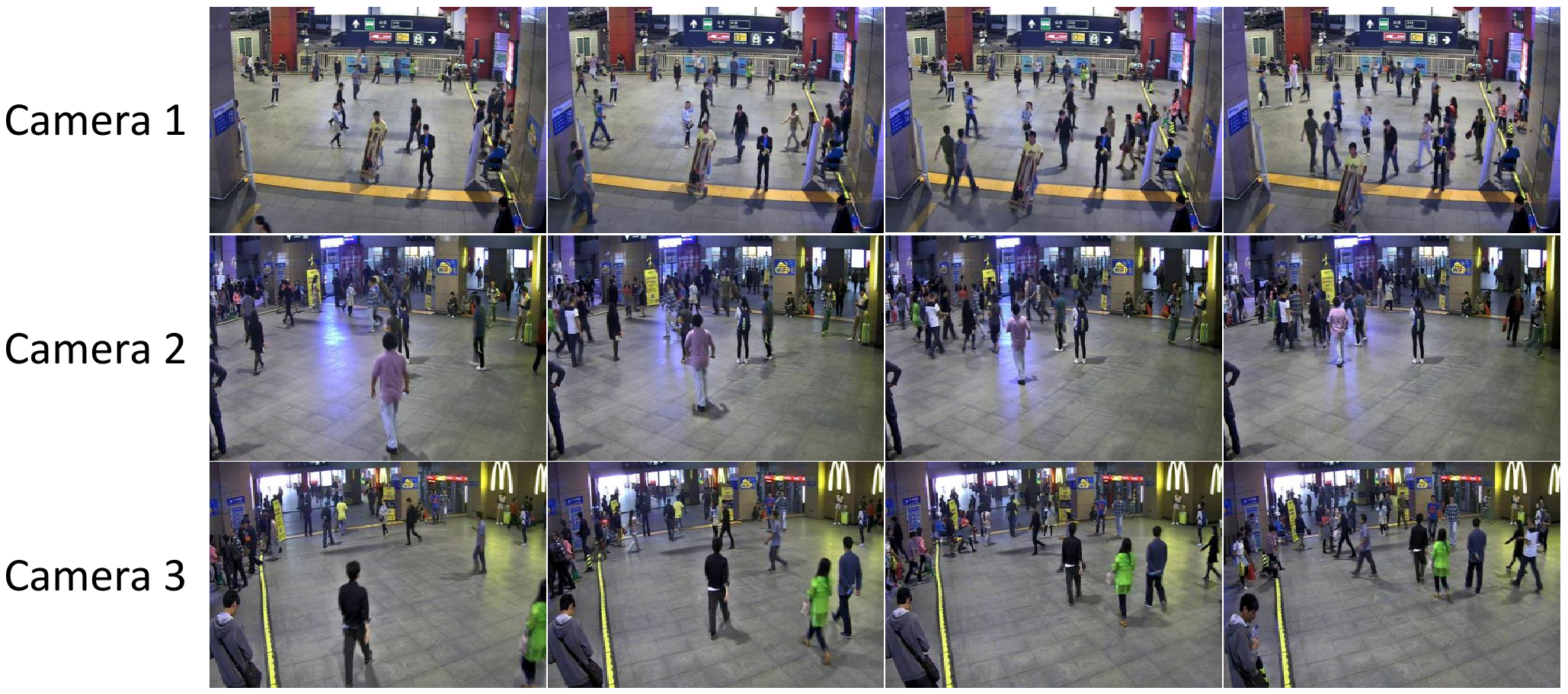}}
\hspace{0.1in}
\subfigure[]{ \label{SKIGfig:c} 
\includegraphics[width=0.87\textwidth]{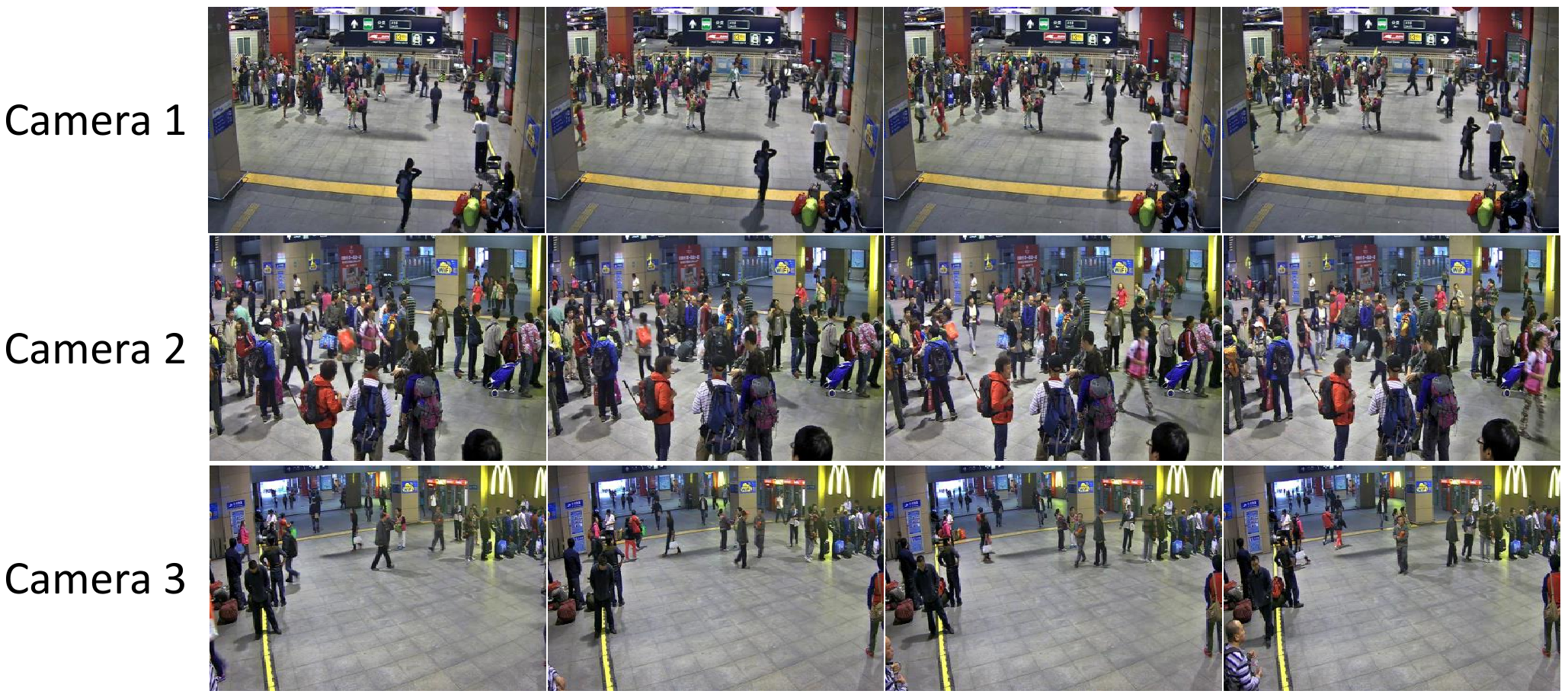}}
\caption{Some samples of three crowd actions in Dongzhimen Transport Hub Crowd dataset. Each row shows frames captured by one of the three cameras. (a) light level. (b) medium level. (c) heavy level.} \label{DZMfig} 
\end{figure*}

\textbf{2) ACT 42 Human Action Dataset}

This dataset consists of 14 complex action patterns performed by 21 subjects, collected from four cameras in different viewpoints. Each type of action is repeated twice by each subject. 
These 14 actions are: ``Collapse'', ``Stumble'', ``Drink'', ``Make phone'', ``Read Book'', ``Mop Floor'', ``Pick up'', ``Throw away'', ``Put on'', ``Take off'', ``Sit on'', ``Sit down'', ``Twist open'', and ``Wipe clean''. Each clip contains 35 to 554 frames. To reduce the computation cost and the memory requirement of all the methods, each image is resized from $480\times 640$ pixels to $32\times 48$. Some frame samples of the ACT 42 dataset are shown in Fig.~\ref{ACTfig}.
\begin{figure*}
    \begin{center}
    \includegraphics[width=0.9\textwidth]{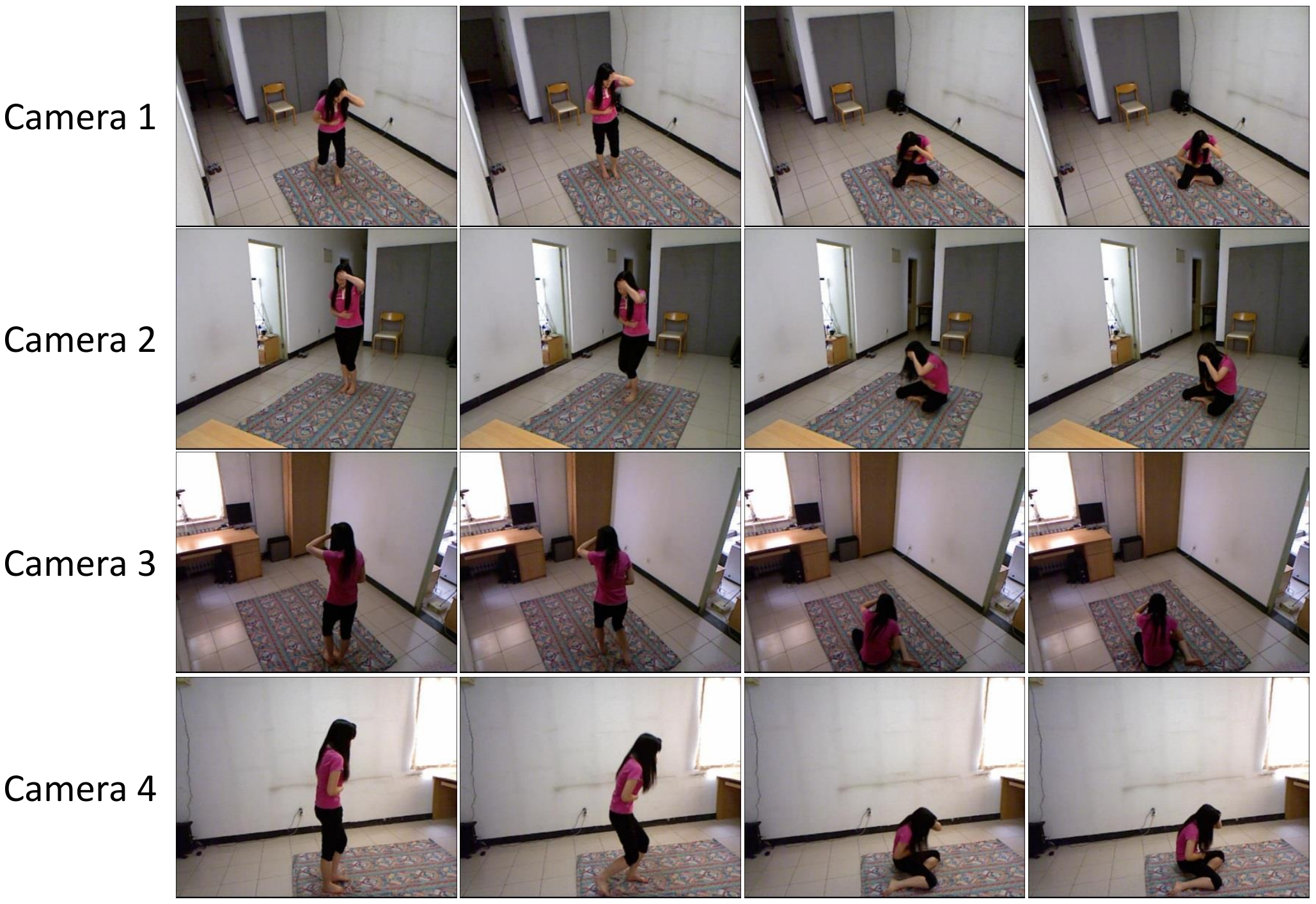}
    \end{center}
    \caption{Some samples in The ACT42 samples. Each row presents a video sequence form a camera. There are 4 cameras to record the same action simultaneously.}\label{ACTfig}
\end{figure*}

\textbf{3) SKIG Action Dataset}

This dataset contains  1080 RGB-D sequences captured by a Kinect sensor. This dataset stores ten kind of gestures of six persons: 'circle', 'triangle', 'up-down', 'right-left', 'wave', 'Z', 'cross', 'come-here', 'turn-around', and 'pat'. All the gestures are performed by fist, finger and elbow respectively under three backgrounds (wooden board, white plain paper and paper with characters) and two illuminations (strong light and poor light). Each RGB-D sequence contains 63 to 605 frames. Here the images are normalized to $24\times 32$ with mean zero and unit variance. Fig. \ref{SKIGfig} shows some RGB and DEPTH images.
\begin{figure*}
    \begin{center}
    \includegraphics[width=0.9\textwidth]{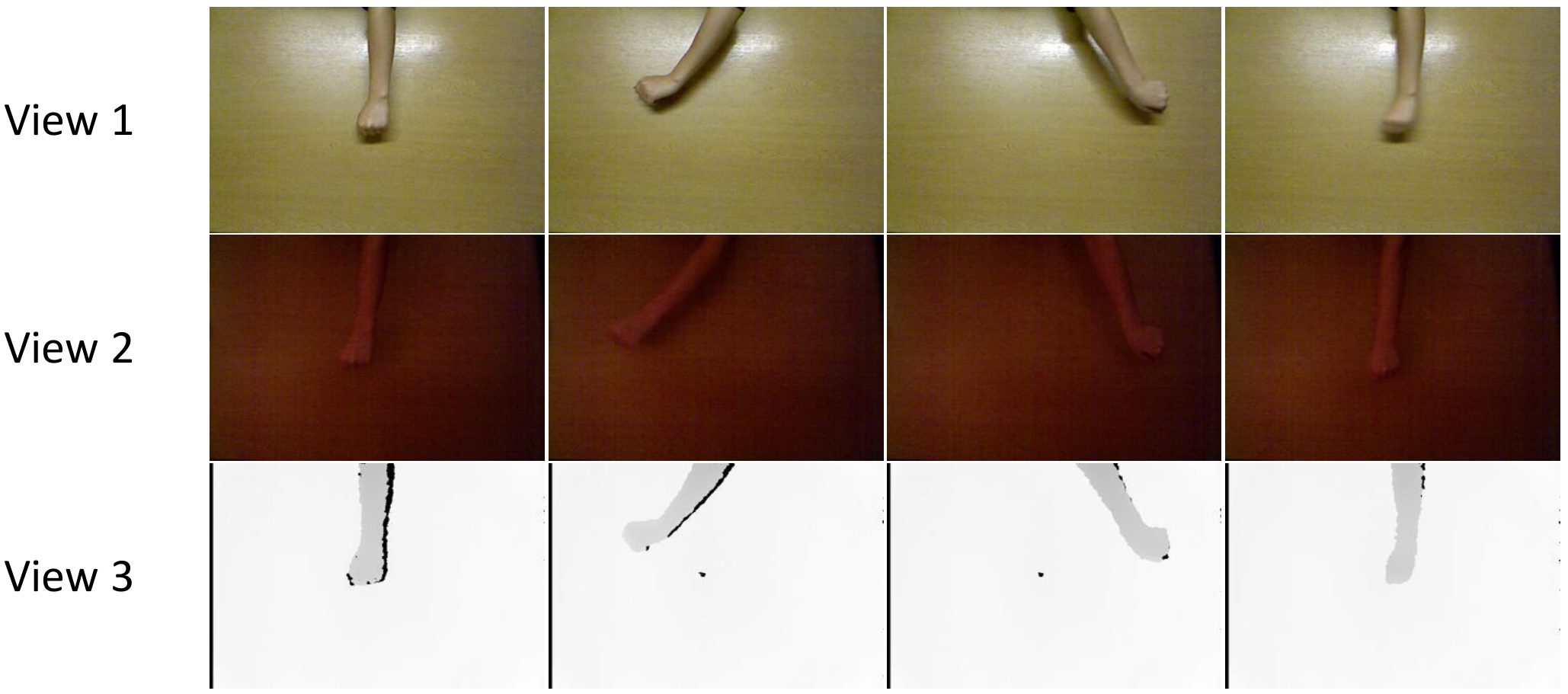}
    \end{center}
    \caption{Some samples with different viewpoints and illuminations in SKIG dataset. View 1 represents light sequences, View 2 represents dark sequences and View 3 represents depth sequences.}\label{SKIGfig}
\end{figure*}

\subsection{Experimental Parameters}
In our experiments, some model parameters in the proposed methods should be adequately adjusted , such as $\lambda$, $\beta$ and $\varepsilon$. To assess the impact of these model parameters, we will conduct experiments by varying one parameter while keeping others fixed to achieve the best parameter values.

$\lambda$ and $\beta$ are the most important penalty parameters for balancing the error term, the low-rank term and Laplacian regularization term in our proposed methods. Empirically, the best value of $\lambda$ depends on  particular applications and has to be chosen from a large range of values to get a better performance. From our experiments, we have observed that when the cluster number is increasing, the best $\lambda$ is decreasing. Additionally, $\lambda$ will be smaller when the noise level in data is lower while $\lambda$ will become larger if the noise level higher. These observations are useful in selecting a proper $\lambda$ value for different datasets. However, the value of $\beta$ is usually very small for various applications with range from $1.0\times 10^{-4}$ to $1.0\times 10^{-2}$. But it does not mean this parameter is unimportant, because the element values of $Z$ is usually thousand smaller than the element values of Laplacian matrix $L$.

The error tolerance $\varepsilon$ is also an important parameter in controlling the terminal condition, which bounds the allowed reconstructed error. We experimentally seek a proper value of $\varepsilon$ to make the iteration process stop at an appropriate level of reconstructed error. Here we set $\varepsilon = 1.0\times 10^{-8}$ for all experiments.


The performances of different algorithms are measured by the following  clustering accuracy
\[
\text{Accuracy} = \frac{\text{number of correctly classified points}}{\text{total number of points}}\times 100\%.
\]

All the algorithms in our experiments are coded in Matlab 2014a and implemented on a machine with Intel Core i7-4770K 3.5GHz CPU and 32G RAM.

\subsection{Dongzhimen Transport Hub Crowd Dataset}
According to Section \ref{Sec:3.1}, for each clip $C_i^m, m=1,2,3$ in the $i$-th sample $[Y_i],i=1,...,182$, we set the subspace dimension $p_m=10$ to construct a Grassmann point $X^m_i\in\mathcal{G}(10,1856),m=1,2,3$. Therefore, we could use a Product Grassmann point  $[X_i]=\{X^1_i,X^2_i,X^3_i\}\in\mathcal{PG}_{1856:10,10,10}, i=1,...,182$ to represent a sample in the dataset.

This is a challenging dataset for clustering, because most video clips contain too much noise or many outliers. For example, most video clips mix up several kinds of crowd actions, for which it is difficult to label. Table \ref{DTHCtab1} presents the clustering results of all the methods.  In this set of experiments, the classic LRR, K-means and SC methods are conducted on the IDT features and LBP-TOP features. Remarkably, our Grassmann manifold-based LRR methods (PGLRR, LapPGLRR and GLRR-F) achieve  much better results, winning at least 1.1 percentage advantage. This demonstrates the Grassmann manifold representation is a better way to represent high dimensional nonlinear data such as multi-camera videos. PGLRR, LapPGLRR and GLRR-F have at least 16.48\% more clustering accuracy than SCGSM. This evidences the effectiveness of LRR too. Our proposed methods, especially GLRR-F, outperform the other methods by at least 4.4\% in accuracy. This fact empirically proves that properly joining multi-camera videos  in terms of product manifolds helps analyzing crowd actions. From this, we conclude that combining LRR, Grassmann manifold representation and product manifold helps improve the clustering accuracy of the model.

We further analyze the functions of three cameras we used.  As showing in Fig. \ref{Fig2}, many actions happened in the yellow area. This  means camera 3  often captures incomplete parts of the actions.  From this observation,  we design another experiment without using the data captured by camera 3.  Thus each Product Grassmann point is constructed by two Grassmann points, i.e. $M=2$. The experimental results are also presented in Table \ref{DTHCtab1}, showing an even better result than that for the case of three cameras. This demonstrates that some unwanted information can degrade the model performance.

\begin{table}
   \centering
   \begin{tabular}{|c|c|c|}
     \hline
              \diagbox{Methods}{Camera Numbers} & 3& 2\\
              \hline
              PGLRR & \textbf{0.8352} & \textbf{0.9176}\\
              \hline
              LapPGLRR & \textbf{0.8352} & \textbf{0.9176}\\
              \hline
              GLRR-F & 0.7912 & 0.7912\\
              \hline
              SCGSM  & 0.6264 & 0.6648\\
              \hline
              SMCE   & 0.6154 & 0.4890\\
              \hline
              LS3C   & 0.4890 & 0.4890\\
              \hline
            LRR+IDT  & 0.5824 & 0.5659\\
              \hline
            K-means+IDT   & 0.6758 & 0.6758\\
              \hline
            SPC+IDT    & 0.4176 & 0.3187\\
              \hline
            LRR+LBP-TOP  & 0.7802 & 0.7802\\
              \hline
            K-means+LBP-TOP  & 0.5275 & 0.7588\\
              \hline
            SPC+LBP-TOP    & 0.4176 & 0.4176\\
     \hline
   \end{tabular}\\[2mm]
  \caption{Subspace clustering results on the Dongzhimen Transport Hub Crowd  dataset.}\label{DTHCtab1}
\end{table}

\subsection{ACT42 Human Action Dataset}

To validate the effectiveness of our proposed methods, we select this dataset which is collected under a relative pure background condition with four cameras in different viewpoints. This dataset has 14 clusters.  Similar to the last experiment setting, we set the subspace dimension $p_m=10$ to construct a Grassmann point $X_i^m\in\mathcal{G}(10,1356)$, $m=1, 2, 3$ and  $4$. Finally, we can obtain totally $588$ Product Grassmann points $[X_i]=\{X_i^1, X_i^2, X_i^3, X_i^4\}\in\mathcal{PG}_{1536:10,10,10,10}$ as the inputs.

This dataset can be regarded as a clean dataset without noises because of controlled internal settings. In addition, each action is recorded by four cameras at the same time and each camera has a clear view, which helps improve the  performance of the evaluated methods. Table \ref{ACTtab1} presents the experimental results of all the algorithms on the ACT42 Human Action dataset.  The three classic clustering methods using both LBP-TOP features and the IDT video features fail to produce satisfactory results (about 11\% lower than PGLRR, LapPGLRR and GLRR-F in accuracy). Once again, this reflects the important role that Grassmann manifold representation plays. As to SCGSM, the gap of 9.19\% confirms LRR makes great contribution to our proposed methods. Meanwhile the proposed method provides better performance than GLRR-F, SCGSM, SMCE, and LS3C by at least 1.5\% in accuracy. This experiment demonstrates the advantage of using the product manifold based representation.

However, the clustering accuracy $0.4745$ is actually bad for 14 clusters in such pure background conditions. The main reason may be due to the fact that too similar actions are contained in this dataset,
such as collapse and stumble, sit on and sit down, etc. To further test the performance of the proposed method on a more meaningful human action dataset, we throw some similar type of action video clips. We only select seven types of actions to create a new dataset for actions ``Collapse'', ``Drink'', ``Mop Floor'', ``Pick up'', ``Put on'', ``Sit up'', and ``Twist open''. The last column in Table \ref{ACTtab1} presents the experimental results of all methods when the cluster number is 7. The proposed method gets the highest accuracy $0.7687$. This shows that the proposed method is suitable to action clustering.
\begin{table}
   \centering
   \begin{tabular}{|c|c|c|}
     \hline
              \diagbox{Methods}{Cluster Numbers} & 14 & 7\\
              \hline
              PGLRR & \textbf{0.4745} & 0.7687\\
              \hline
              LapPGLRR & 0.4728 & \textbf{0.7823}\\
              \hline
              GLRR-F & 0.4575 & 0.6701\\
              \hline
              SCGSM  & 0.3656 & 0.4966\\
              \hline
              SMCE   & 0.4507 & 0.5748\\
              \hline
              LS3C   & 0.4422 & 0.5442\\
               \hline
             LRR+IDT & 0.3425 & 0.3374  \\
              \hline
            K-means+IDT & 0.3333 & 0.4479  \\
              \hline
            SPC+IDT  & 0.1446 & 0.2883  \\
            \hline
            LRR+LBP-TOP & 0.1446 & 0.2483  \\
              \hline
           K-means+LBP-TOP & 0.1327 & 0.2177  \\
              \hline
            SPC+LBP-TOP  & 0.1429 & 0.1429  \\
     \hline
   \end{tabular}\\[2mm]
  \caption{Subspace clustering results on the ACT42 Human Action dataset.}\label{ACTtab1}
\end{table}

\subsection{SKIG Action Dataset}
The video clips in this dataset contain illumination variety and background variety. Generally, the number of Grassmann manifolds in the product space is determined by the number of varying factors existed in data. Here, the main underlying factors are light illumination, dark illumination, and depth. As there are many varying factors for one kind of gesture here, we design different types of Product Grassmann points with different combinations of factors, including:
light + depth sequences $([X_i]=\{X_i^1,X_i^2\}\in\mathcal{PG}_{1024:20,20})$;
light + dark  sequences $([X_i]=\{X_i^1,X_i^2\}\in\mathcal{PG}_{1024:20,20})$; 
dark  + depth sequences $([X_i]=\{X_i^1,X_i^2\}\in\mathcal{PG}_{1024:20,20})$; and 
light + dark  + depth sequences $([X_i]=\{X_i^1,X_i^2,X_i^3\}\in\mathcal{PG}_{1024:20,20,20})$.
In the clustering experiment, for each Product Grassmann Manifold type, we select 54 samples from each of ten clusters.

\begin{table*}
   \centering
   \begin{tabular}{|c|c|c|c|c|}
     \hline
              \diagbox{Methods}{Data Type} & light+depth & light+dark  & dark+depth & light+dark+depth\\
              \hline
              PGLRR & 0.5907 & 0.6000 & 0.6833 & 0.6315\\
              \hline
              LapPGLRR & \textbf{0.6537} & \textbf{0.6870} & \textbf{0.6981} & \textbf{0.6685}\\
              \hline
              GLRR-F& 0.5685 & 0.5185 & 0.6148 & 0.5944\\
              \hline
              SCGSM & 0.4093 & 0.4667 & 0.5056 & 0.4296\\
              \hline
              SMCE  & 0.4481 & 0.4130 & 0.6389 & 0.5796\\
              \hline
              LS3C  & 0.4907 & 0.3722 & 0.6333 & 0.5833\\
               \hline
              LRR+IDT  & 0.5463 & 0.5963 & 0.6019 & 0.5963\\
              \hline
              K-means+IDT   & 0.4685 & 0.4759 & 0.6426 & 0.5407\\
              \hline
             SPC+IDT   & 0.1000 & 0.2000 & 0.2000 & 0.4000\\
             \hline
             LRR+LBP-TOP  & 0.222 & 0.2167 & 0.2352 & 0.2056\\
              \hline
             K-means+LBP-TOP   & 0.1704 & 0.1444 & 0.1870 & 0.1626\\
              \hline
             SPC+LBP-TOP   & 0.1000 & 0.2000 & 0.1000 & 0.1009\\
     \hline
   \end{tabular}\\[2mm]
  \caption{Subspace clustering results on the SKIG dataset.}\label{SKIGtab}
\end{table*}

In this experiment, we want to study how to select proper  views to obtain the best clustering accuracy. From Table \ref{SKIGtab}, we find an interesting phenomenon that the experimental result for the case of dark+depth is obviously better than light+dark+depth. The difference is the dark video clips in these two conditions. We believe the outline of an object will be clear and the background will fade when the illumination becomes darker. Similarly, the outline of an object will be fused with the background as the illumination becomes lighter. This condition may decrease the clustering accuracy. It is well-known that depth camera can extract human skeletons as well, which is very meaningful for action clustering. Thus, to obtain higher clustering accuracy, we need to analyze the function of each camera or each type of data. Similar to previous experimental results, we note that the proposed methods, PGLRR and LapPGLRR, are obviously superior to other methods. We contribute this to the advantages of the product manifold, Grassmann manifold representation and LRR for multi-camera videos.

\section{Conclusion}\label{Sec:6}
In this paper, we propose a data representation method based on the Product Grassmann manifold for multi-camera video data. By exploiting the geometry metric on the product manifold, the LRR based subspace clustering method is extended to obtain an LRR model on the Product Grassmann manifold.  An efficient algorithm is also proposed for the new model. In addition, we introduce the Laplacian constraint into the new LRR model on the Product Grassmann Manifold. The high performance in the clustering experiments on different video datasets indicates that the new model is well suitable for representing non-linear high dimensional data and revealing intrinsic multiple subspaces structures underlying data. In the future work, we will focus on investigating different metrics of the Product Grassmann Manifold and test the proposed methods on large scale complex multi-camera videos.

\section*{Acknowledgements}
The research project is supported by the Australian Research Council (ARC) through grant DP140102270 and also partially supported by the National Natural Science Foundation of China under Grant No. 61390510, 61133003, 61370119 and  61227004, Beijing Natural Science Foundation No. 4132013 and 4162010, Project of Beijing Educational Committee grant No. KM201510005025 and Funding PHR-IHLB of Beijing Municipality.


\bibliographystyle{IEEEtran}

\begin{IEEEbiography}[{\includegraphics[width=1in,height=1.25in,clip,keepaspectratio]{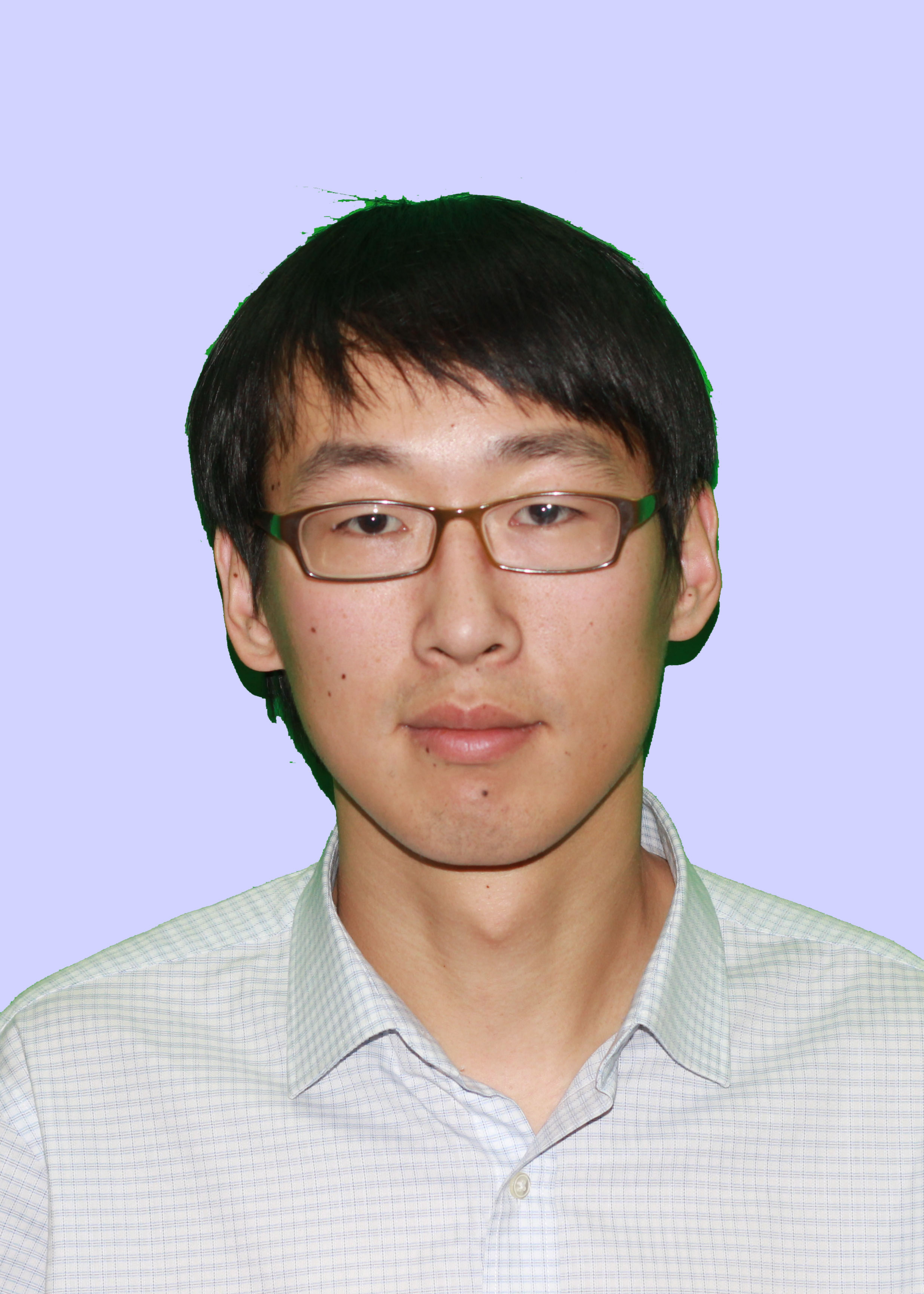}}]
{Boyue Wang} received the B.Sc. degree from Hebei University of Technology,
Tianjin, China, in 2012. he is currently pursuing the
Ph.D. degree in the Beijing Municipal Key Laboratory of Multimedia and Intelligent Software Technology,
Beijing University of Technology, Beijing.
His current research interests include computer
vision, pattern recognition, manifold learning and kernel methods.
\end{IEEEbiography}

\begin{IEEEbiography}[{\includegraphics[width=1in,height=1.25in,clip,keepaspectratio]{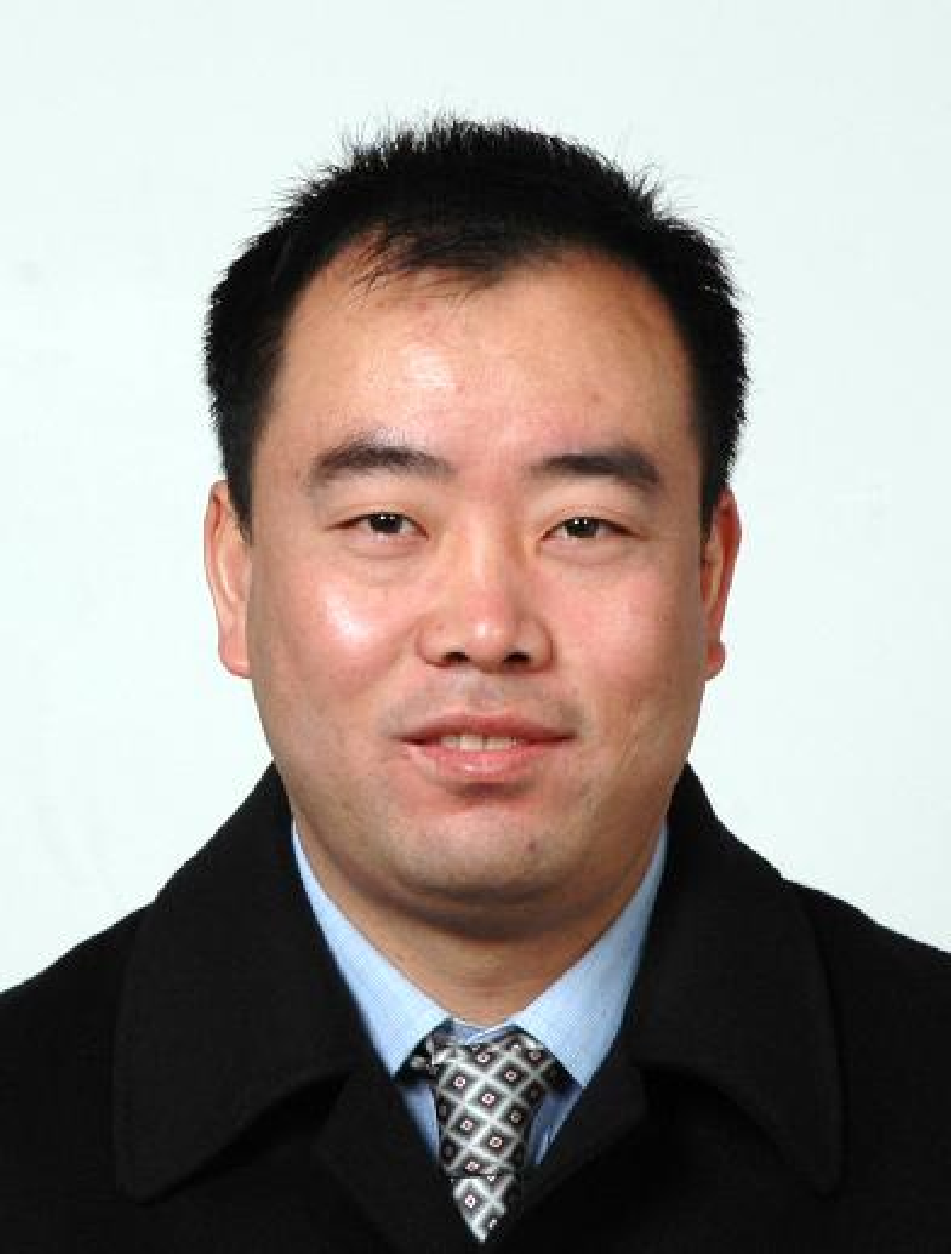}}]
{Yongli Hu} received his Ph.D. degree from Beijing University of Technology in 2005. He is a professor in College of Metropolitan Transportation at Beijing University of Technology. He is
a researcher at the Beijing Municipal Key Laboratory of Multimedia and Intelligent Software Technology.
His research interests include computer graphics, pattern recognition and multimedia technology.
\end{IEEEbiography}

\begin{IEEEbiography}[{\includegraphics[width=1in,height=1.25in,clip,keepaspectratio]{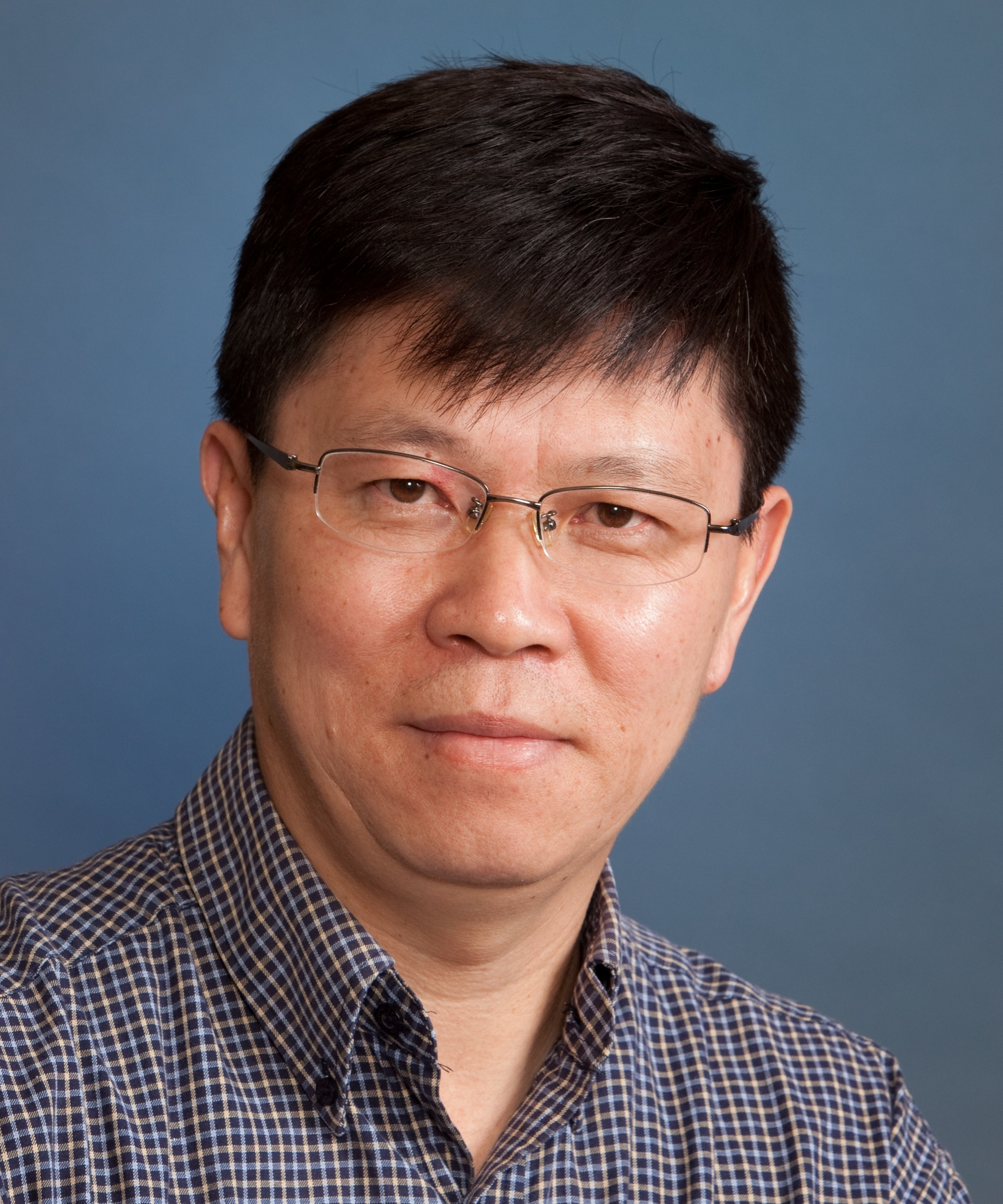}}]
{Junbin Gao} graduated from Huazhong University of Science and Technology (HUST),
China in 1982 with BSc. degree in Computational Mathematics and
obtained PhD from Dalian University of Technology, China in 1991. He is the  Professor of Big Data Analytics in the University of Sydney Business School at the University of Sydney and was a Professor in Computer Science
in the School of Computing and Mathematics at Charles Sturt
University, Australia. He was a senior lecturer, a lecturer in Computer Science from 2001 to 2005 at
University of New England, Australia. From 1982 to 2001 he was an
associate lecturer, lecturer, associate professor and professor in
Department of Mathematics at HUST. His main research interests
include machine learning, data analytics, Bayesian learning and
inference, and image analysis.
\end{IEEEbiography}

\begin{IEEEbiography}[{\includegraphics[width=1in,height=1.25in,clip,keepaspectratio]{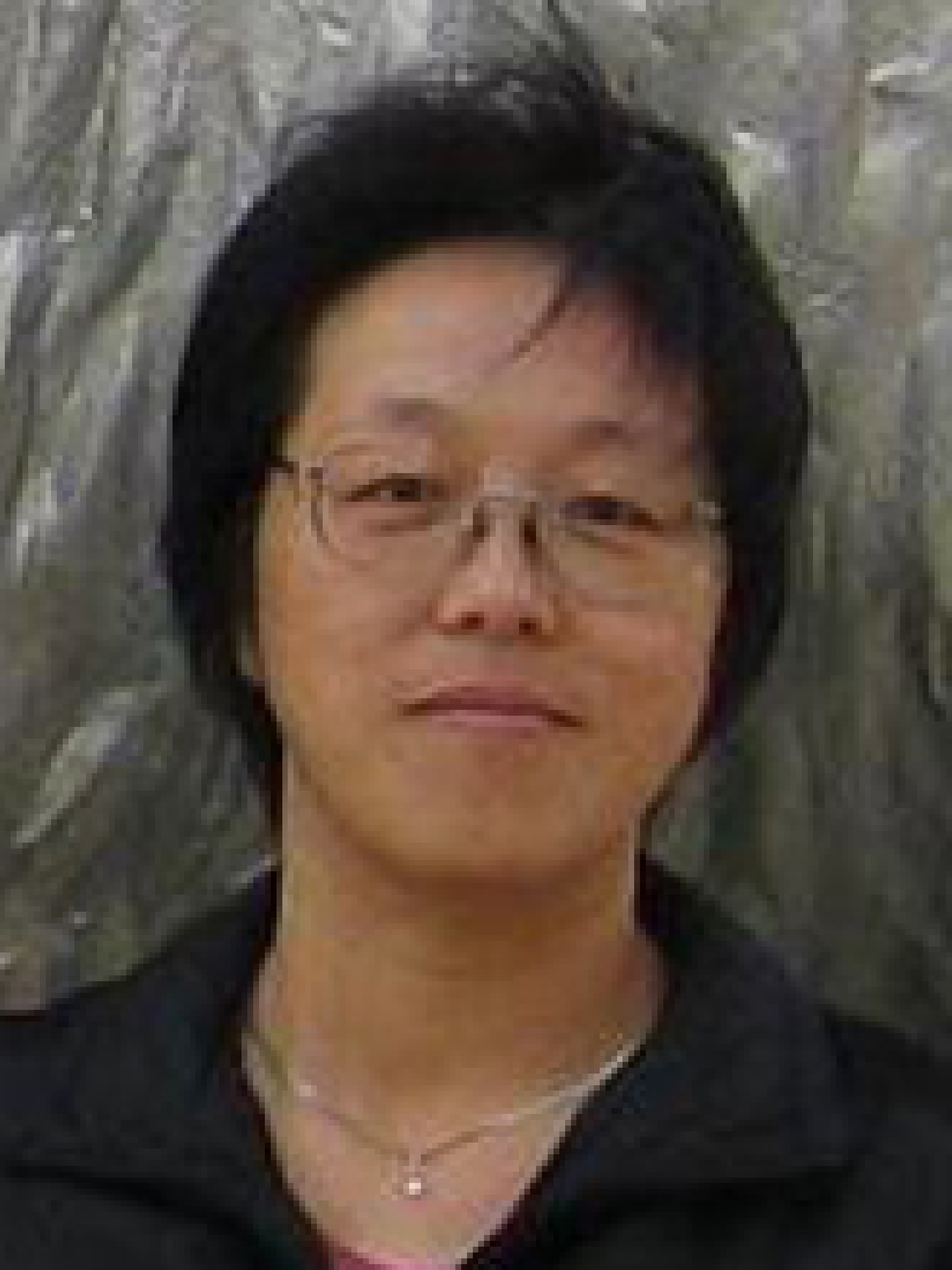}}]
{Yanfeng Sun} received her Ph.D. degree from Dalian University of Technology in 1993. She is a professor in College of Metropolitan Transportation at Beijing University of Technology. She is
a researcher at the Beijing Municipal Key Laboratory of Multimedia and Intelligent Software Technology. She is the membership of China Computer Federation.
 Her research interests are multi-functional perception and image processing.
\end{IEEEbiography}

\begin{IEEEbiography}[{\includegraphics[width=1in,height=1.25in,clip,keepaspectratio]{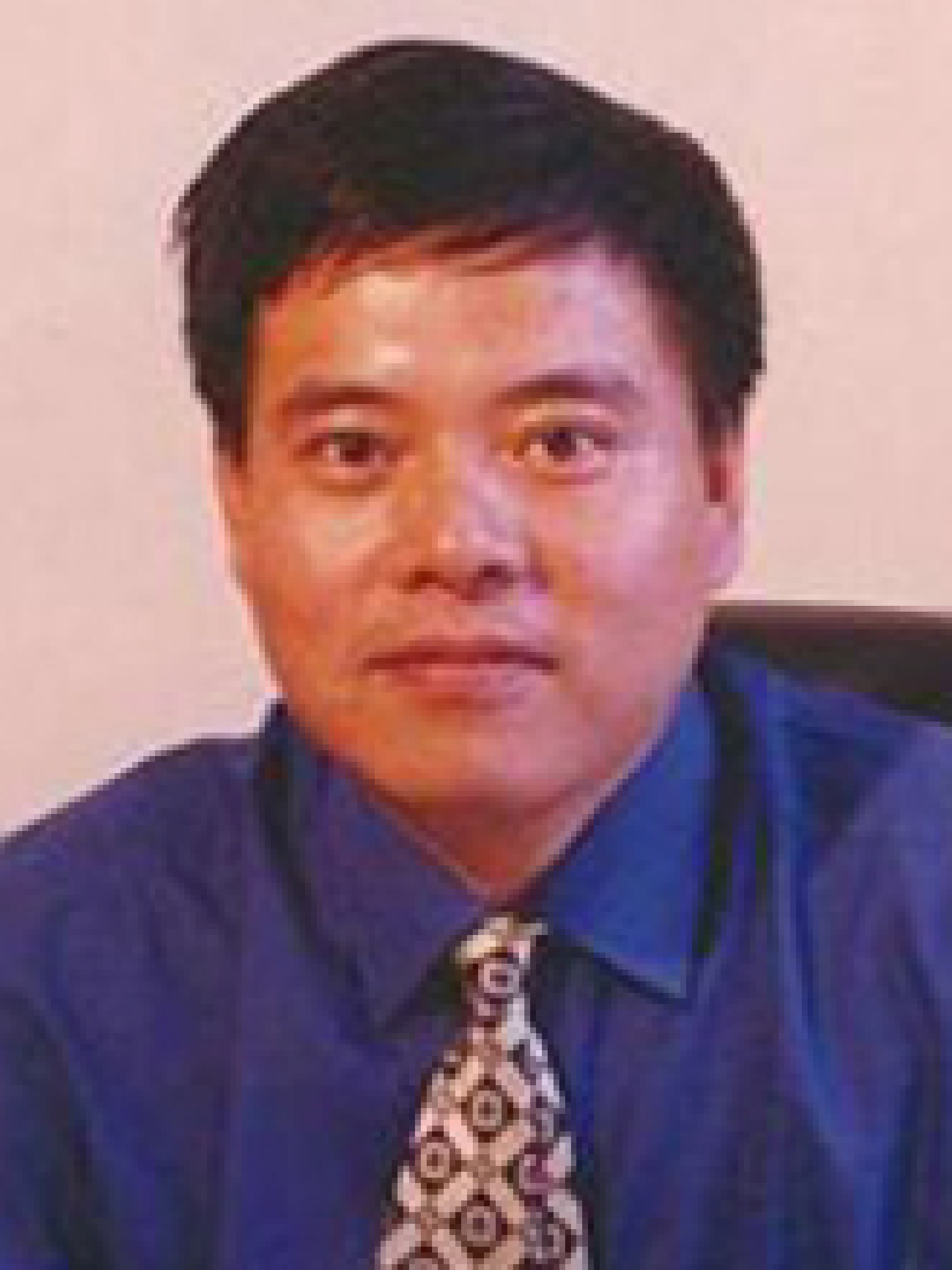}}]
{Baocai Yin} received his Ph.D. degree from Dalian University of Technology in 1993.
He is a Professor in the College of Computer Science and Technology, Faculty of Electronic Information and Electrical Engineering, Dalian University of Technology. He is
a researcher at the Beijing Municipal Key Laboratory of Multimedia and Intelligent Software Technology.
He is a member of China Computer Federation. His
research interests cover multimedia, multifunctional perception, virtual reality and computer graphics.
\end{IEEEbiography}
\vfill

\end{document}